\documentclass{article} 
\usepackage{aaai18}  
\usepackage{times}  
\usepackage{helvet}  
\usepackage{courier}  
\usepackage{url}  
\usepackage{graphicx}  
\usepackage{amsmath,amssymb,amsfonts}
\nocopyright
\begin{document}
%
\title{Window detection in aerial texture images of the 3D CityGML Berlin Model}
\author{Franziska Lippoldt\\
Fraunhofer IDM \\ at Nanyang Technological University\\
Singapore \\
flippoldt@ntu.edu.sg \\
\And 
Marius Erdt \\
Nanyang Technoligcal University \\ and Fraunhofer Singapore \\
Singapore\\
merdt@ntu.edu.sg
}
\maketitle

\begin{abstract}
This article explores the usage of the state-of-art neural network Mask R-CNN to be used for window detection of texture files from the CityGML model of Berlin.\\
As texture files are very irregular in terms of size, exposure settings and orientation, we use several parameter optimisation methods to improve the precision. Those textures are cropped from aerial photos, which implies that the angle of the facade, the exposure as well as contrast are calibrated towards the mean and not towards the single facade. The analysis of a single texture image with the human eye itself is challenging: A combination of window and facade estimation and perspective analysis is necessary in order to determine the facades and windows.\\
We train and detect bounding boxes and masks from two data sets with image size 128 and 256. We explore various configuration optimisation methods and the relation of the Region Proposal Network, detected ROIs and the mask output.
Our final results shows that the we can improve the average precision scores for both data set sizes, yet the initial AP score varies and leads to different resulting scores.
\end{abstract}

\section{Introduction}

\begin{figure*}[h]
	\centering
	\includegraphics[width=\textwidth]{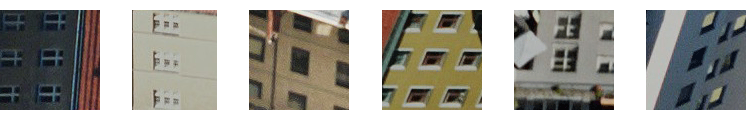}
	\caption{Sample windows from textures}\label{windows}
	
\end{figure*}

A complete list of 26 international cities has been publishing their city 3D models online to be available for public, most of them available in CityGML format as established in \cite{groger2012citygml}. While several of those also contain textures, the 3D city models have a simplified shape: in CityGML terms they are at LOD2, providing the basic shape of the build but omitting facade details. \\
The idea of window detection and facade segmentation originates from the idea of automatically integrating the 3D features into the 3D models. In order to do so, we need to analyze the given texture images through image segmentation.\\
Due to the strong correspondence of the texture files of the Berlin 3D model of \cite{dollner2006virtual} with satellite images, the results of this paper can also be applied to satellite or drone captures. This is especially interesting since satellite images are provided in super resolution and need to be cropped into suitable parts. In contrast to those "selected crops", the crops of the CityGML model are determined in advance and cannot be rearranged to a whole image.\\
In order to segment the facade and analyze the windows, we propose to use a semantic neural network. The network's main components are a masking part of the facade and a facade segmentation.
While this structure has been deployed in the Mask R-CNN \cite{he2017mask} originally to detected humans and deliver fast region proposal for semantic segmentation, we shift the usage to a very different area of application. The original "mask" was deployed on the image to suggest regions faster and more accurate object detection, we use the "mask" to build a facade aware segmentation to find windows.

\section{Related Work}
Since 2015, semantic image segmentation has been tackled as one of the main parts of neural networks. Segnet \cite{badrinarayanan2015segnet} was one of the first to tackle the problem of semantic segmentation of the whole image. Up to this year, several implementations for image segmentation exist. Recent challenges of this year deal with the semantic segmentation of satellite data. The winning networks of the satellite changes were mostly combined neural networks, which assign different task to different neural networks in pipelines. In general, image segmentation tasks can be more accurate when separately training networks for the objects and then combining those networks together, see \cite{lin2016efficient}.\\
As an applied example of semantic segmentation, facade segmentation has been studied in several works. Facade segmentation from street view style photos has been reduced to the task of finding repetitive objects or grid structures on the facade, depending on the architectural style.
Major works were done by \cite{mathias2016atlas} and \cite{liu2017deepfacade}.
\\
More recently this year, the DeepGlobe challenge \cite{DeepGlobe18}, a set of three contests for satellite data object recognition, has been started. Whereas segmentation methods such as the SegNet proposed a single network with one training data set, those recent DeepGlobe solutions offer higher flexibility and directed training towards weak points of the object recognition.\\
The current state-of-the-art neural network for semantic segmentation for detection of the exact shape of humans is the Mask R-CNN, which shows both improvements in speed as well as a very good accuracy. Its main components are a Region Proposal Network (RPN) for recommending appropriate regions in the image and the main detection and segmentation network, which segments the image into objects pixel-wise. It is the extension of Fast R-CNN \cite{girshick2015fast} and Faster R-CNN \cite{ren2015faster}, which have been focusing on region detection enhancement.

\section{Problem Analysis}
\subsection{Texture image analysis}
The texture image files show the following variations:
\begin{enumerate}
	\item Varying exposure settings
	\item Perspective distortion
	\item Partially covering Shadows
\end{enumerate}
While technically perspective distortion can be calculated and inverted, it is very tricky to do so in combination with varying exposure and partial shadows. In Figure \ref{windows}, a subset of windows is displayed. For highly underexposed or overexposed images, the border of the windows is lost. 
All together, those issues pose already significant problems for human eyes to spot the windows. The analysis of possible window candidates can only be confirmed by detecting features on the image, i.e. rooftop tiles or facades. While facades are basically a grid of windows, windows are part of the facade. 
While it is clear that the set of all windows of a house define the facade and the facade contains the set of all windows, the given input images will leave neither one or the either completely resolved.


\subsection{Training data}
In the process of optimizing the trained network, we constantly adjusted and modified the training set. Two different data sets of images of fixed size have been labeled and been used for experimenting with the neural network. While the texture image size is between 100 and 300 px, we chose to select images which can be cropped to size 128 and 256 respectively.\\
\textbf{Data set 128}
The first data set contains crops of the texture images of size 128x128. We have labeled the whole texture file and then cropped them adaptively such that each cropped image has a maximal overlay of $10\%$ with the previous one. We normalized those results with respect to the histogram. Further augmentation has been done in terms of $90 \deg$rotations without loss of quality. 
We have randomly shuffled the images and divided the dataset into training, validation and testing with ratio 6:2:2.\\
Overall this data set contains around 6,000 images.\\
\textbf{Data set 256}
The second data set contains adaptive crops of hand-labeled images of size 256x256. We have chosen the best texture files from a set of more than 1,000 images in terms of resolution, image dimensions and exposure settings. Most of those chosen images are not much larger in terms of side length than 300px. While we performed the same adaptive cropping and augmentation as in the 128 data set, the resulting training and evaluation set is slightly smaller in terms of number of images. We divided the augmented images randomly into training, validation and test sets the same way as the other data set.\\
The data set contains 1,000 images.

\section{Optimization methods}
In order to modify and optimize the given neural network, we analyzed critical features and relations of those in three separate steps, according to what point in time in the process they need to be set up and on which level of modification they can invoke. The optimization of the structure of the network is the very 'low level' approach to optimise the correctness of the results, while training optimisation is done in context of the dataset and inference optimisation adjust the outcome of the inference model.

\subsection{Structure optimization}
Assume to start with the structure of the Mask R-CNN as described in the paper \cite{he2017mask} and the source code provided in \cite{matterport_maskrcnn_2017}.
We analyse that the main Mask R-CNN has five downsampling and five upsampling units. It works with an input image of size nxn with n fixed as 1024. The Region proposal network used for the recommendation of the mask consists of 5 downsampling units. Each of those downsampling units is decreasing the image size by a factor of two, while the features are increased by a factor of two. So to say, the product of both is constant. \\
We will use the Mask R-CNN in combination with the ResNet101 Backbone as described in \cite{he2016deep}. The backbone and its five stages are used to make the region proposal of the Region Proposal Network (RPN). The resulting regions of interest (ROIs) are then aligned to reduce the error when using convolutions with a stride 2 or 3. Afterwards, the head of the network, a six stage Feature Pyramid Network (FPN) is used to predict the mask and a five stage FPN to predict bounding boxes and classes.
In this context, we are very aware that the main application of the Mask R-CNN is for the detection of people on a 1024px square image. Facade images as given in the Berlin CityGML model are essentially different in size, quality and resolution. They are around four to six times smaller. While the standard Mask R-CNN network would automatically enlarge those photos, we do realize that this is not an optimal setting. To use an input that is smaller by a factor of 5, while using the same amount of features and the same layer sizes, would automatically lead to over fitting. Even though one might argue that more images can reduce the problem of over fitting, the amount of images cannot change the amount of information available on those images, i.e. the dimensions. As a result of this, we radically modified the structure and adjusted the dimensions using the following model: \\
Without loss of generality, let us assume that we have a dataset of k images $img_1,\dots ,  img_k$ with standard input size 1024 and  width a 
\begin{equation*}
w(\text{img}_i) = \dfrac{1024}{2^{m}} \text{ ,where }  m \in \{ 2,3,4,5\}
\end{equation*}.

Let $k_{\text{layer}}$ be the number of convolutional stages with stride two. We know that the height and width of the image in the lowest layer then equals:
\begin{equation*}
\dfrac{w(\text{img}_i) }{2^{k_{\text{layer}}}} = \dfrac{1024}{2^{m k_{\text{layer}}}} 
\end{equation*}. \\ 
In particular, assume that the average object size to be detected on the images is around $10\%$ of the whole image size: $w(o_{\text{img}_i}) <  10 \% w(\text{img}_i) $.  Then the object size in the lowest layer is then smaller then 10 Percent of the image size in the lowest layer: 
\begin{equation} \label{eq:window_width_est}
w(o_{{\text{img}}_i})  < \dfrac{ 0.1 * 1024}{2^{k_{\text{layer}}m}} 
\end{equation} 
In this context the optimal depth of the down sampling layers is the L such that the object has appropriate size $p$ with $p > 3$.
We look for a new network structure that is dependent on the data input and specifically the object size after all downsampling layers.
For the given Loss function 
\begin{equation}\label{eq:loss}
L = \alpha L_1 + \beta L_2 + \gamma L_3 + \delta L_4 + \epsilon L_5 ,
\end{equation}
where the first two losses are the RPN losses for the class and bounding box prediction respectively and the other three  losses those of the head neural networks classes, bounding box and mask loss.
We are looking for $k_{layer}$ and corresponding configuration $C_{k_{layer}}$ such that it minimizes the overall loss:
\begin{equation*}
\min_{C_{k_{layer}}} L
\end{equation*}
\subsection{Training optimization}
Training optimization mainly refers to the adaptive process of matching the results of the loss function with the data set and parameters as a result of a trial process. Given a loss $L_i$ and equation \ref{eq:loss}, optimise $K  = (\alpha ,\beta ,\gamma ,\delta,\epsilon) \in \mathbf{R}^{4}$ and $L_1,L_2,L_3, L_4, L_5$ Loss functions such that the average precision for threshold 0.5 $AP_{50}$ of the network is improved, i.e. :
\begin{equation*}
\max_{K , L_1, L_2, L_3, L_4,L_5} AP_{50}
\end{equation*}
The choice of parameters chosen depends on the structure configurations obtained in the previous section and therefore also on the data input.
While technically for $K$ any combination of real values can be combined, think of $K$ as an equivalence class with 
\begin{equation}
(\alpha , \beta , \gamma , \delta,\epsilon ) \sim  n \cdot (\alpha , \beta , \gamma , \delta , \epsilon) , \forall n \in \mathbb{R}
\end{equation}
Working with equivalence classes for optimisation works because even though the total loss value changes, the accuracy of the detection is not dependant on the loss value but the loss gradient. \\
Further more, several other configuration parameters need to be tested and changed to improve the accuracy. The overall success of the training also depends on the amount of objects per image and corresponding region recommendation and anchor settings.
\subsection{Inference optimization}
After training and cross validation have been completed, we save the models parameters. The output of the interference is a list of possible detection boxes and masks, whose probabilities are larger then a threshold $p_{\min}$. Depending on the properties of the images in the test set $data_{test}$, this threshold needs to be optimised with respect to the overall accuracy of the test dataset. \\
If the test images share similar properties with the training images, we can expect an accurate outcome. Even testing images of a completely different size can become increasingly hard, as the ressemblence of similarity is given for the human eye but not necessarily for the neural network working with matrixes of pixels.

\section{Results}

We have run more than 30 configurations of the neural network. Out of those, we chose the best five parameter configurations for each data set. The following results are discussed on the training of two epochs and fine tuning of 3 epochs, where one epoch contains all training set data and validation steps are set to the number of validation images.\\
Because more training might still improve the overall scores, we need to consider the following results to decide which parameters are the right choice for optimization and which kind of optimization improvements we can expect in general. 

\begin{table}[ht]
	\begin{center}
		\caption{Models and scores}
		\label{compare:nn}
		\begin{tabular}{l*{4}{c}r}
			Configurations and Test         & Recall & Precision & $AP_{50}$   \\
			\hline
			A. 128 - standard            & 0.53 & 0.85 & 0.85  \\
			B. 128 - optimised              & 0.60 & 0.82 & 0.87   \\
			C. 256 - standard                & 0.51 & 0.94 & 0.91  \\
			D. 256 - optimised                  & 0.58 & 0.90 & 0.93  \\
			
		\end{tabular}
	\end{center}
\end{table}

\subsection{Recall and precision analysis}
In order to use the networks inference results for detecting windows and analysing the architecture of a facade, we are genuinely interested in detecting every existing window as well as reducing the number of wrongly detected windows. \\
In other words, the recall is a crucial value to improve. As can be seen in Table \ref{compare:nn}, the optimisation of the AP goes hand in hand with the optimisation of the recall. However, we have also realised that for any type of recall and AP improvement, the precision value has slightly decreased by at least 0.02, see Table \ref{compare_delta:nn}. In none of the AP improved configurations have we seen an improvement of the precision, however the recall has always been higher by at least 0.01 and at most 0.07.  
\begin{table}[!t]
	\begin{center}
		\caption{Overview of improvements made through optimisation}
		\label{compare_delta:nn}
		\begin{tabular}{l*{4}{c}r}
			Data set         &  $\Delta$ Recall &  $\Delta$Precision &  $\Delta$ $AP_{50}$  \\
			\hline
			A. 128             & 0.07 & -0.02 & 0.02  \\
			B. 256                & 0.07 & -0.04 & 0.02  \\			
		\end{tabular}
	\end{center}
\end{table}

During the configuration testing, we have tried to increase the precision by changing the loss function parameters $(\alpha , \beta , \gamma , \delta,\epsilon ) $ in equation \ref{eq:loss} such that the mask score is three times higher than the other scores. However, this approach did not improve the precision. We conclude that the precision of the mask results depends strongly on the complete structure of the network and each layer and feature size. We think that the lack of information in the image in terms of image size and resolution leads to a restriction in the possible features and hence a threshold for the possible precision values.

\subsection{Image size comparisons}
Because we have been able to test the network on two different data sets with two different images sizes, we have realized that there is an essential difference of score results for those two. Using the standard configurations without optimisation, we can see that the resulting APs differ by a value of 0.06. The improvements made through parameter optimisation however are similar. This leads to a AP score of 0.87 for the 128 data set and 0.93 for the 256 data set respectively.\\
We have come to the conclusion that the image size is dependent on the overall score of the $AP_{50}$, but still improvements could be made for both data sets in a very similar range.

\begin{figure}[h]
	\centering
	\includegraphics[width= 0.8\linewidth]{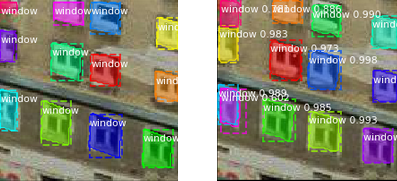}
	\caption{Window detection result example for double window detection}
	\label{fig:windows_crowded}
\end{figure}

\subsection{Anchors, ROIs and AP scores }
\begin{figure}[!ht]
	\centering
	\includegraphics[width= 0.8\linewidth]{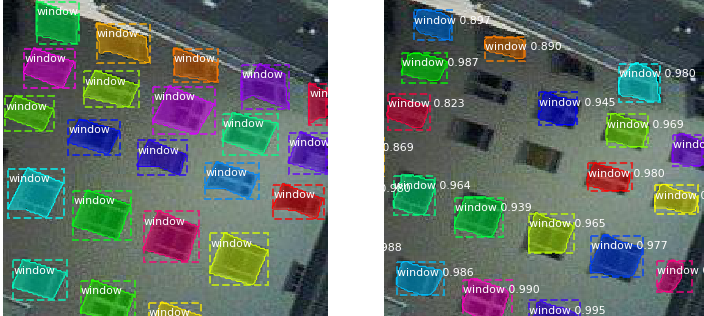}
	\caption{Window detection results for under fitting}
	\label{fig:windows_missing}
\end{figure}

Due to the small image and object size, the anchor size of region proposal in the neural network has been modified in most of the configurations. We have chosen significantly smaller anchor scales and different anchor ratios.\\
Another very crucial parameter to the improvement of the $AP_{50}$ score is the number of trained regions per image. We found that adjusting the number of trained regions per image to reflect the number of windows per image improves the recall values. If chosen improperly, several windows are either not detected at all or windows are double detected, i.e. several regions span over one window and classify it as three objects. An example of missing windows is shown in Figure \ref{fig:windows_missing} and example of double regions is shown in Figure \ref{fig:windows_crowded}. \\
Regarding the RPN, the number of regions for training showed a correlation to the recall value and the under and over fit of the network with respect to a single image evaluation. In our test case, the missing windows often occur towards the center of the image and the dimension of the window as described in equation \ref{eq:window_width_est} were not indifferent from the other detected windows.

\section{ Acknowledgments}
This research is supported by the National Research Foundation, Prime Minister’s Office, Singapore under its International Research Centres in Singapore Funding Initiative.\\
We thank Henry Johan for his support. 
We would also like to thank Rolf Versluis for the additional hardware support, as well as 
Ariya Priyasantha for his technical guidance.\\

\bibliographystyle{ACM-Reference-Format}
\bibliography{paper}

\end{document}